\documentclass[journal]{IEEEtran}
\usepackage{latexsym,bm,amsmath,amssymb}
\usepackage{enumerate}
\usepackage{graphicx}
\usepackage{epstopdf}
\usepackage{subfigure}
\usepackage[ruled,vlined]{algorithm2e}
\usepackage{algorithmic}
\usepackage{amsfonts,amssymb,amsmath}
\usepackage{dsfont}
\usepackage{multirow}
\usepackage{bigstrut,bigdelim,multirow}
\usepackage[table]{xcolor}
\usepackage{colortbl}
\usepackage[T1]{fontenc}
\usepackage{bm}
\usepackage{setspace}
\usepackage{color}
\usepackage{cite}

\newcommand{\tabincell}[2]{\begin{tabular}{@{}#1@{}}#2\end{tabular}}

\graphicspath{{figures/}}

\begin{document}

\title{Biologically inspired model simulating \\visual pathways and cerebellum function in human\\
--\begin{LARGE} Achieving visuomotor coordination and high precision movement with learning ability\end{LARGE}}

\author{Wei~Wu,~\IEEEmembership{Member,~IEEE,}
        Hong~Qiao,~\IEEEmembership{Senior~Member,~IEEE,}
        Jiahao~Chen,
        Peijie~Yin,
        and~Yinlin~Li
\thanks{W. Wu is with the State Key Lab of Management
and Control for Complex Systems, Institute of Automation, Chinese Academy of Sciences, Beijing 100190, China} 

\thanks{H. Qiao is with the State Key Lab of Management
and Control for Complex Systems, Institute of Automation, Chinese Academy of Sciences, Beijing 100190, China and CAS Center for Excellence in Brain Science and Intelligence Technology (CEBSIT), Shanghai 200031, China (e-mail: hong.qiao@ia.ac.cn).}

\thanks{J. H. Chen is with the State Key Lab of Management and Control for Complex Systems, Institute of Automation, Chinese Academy of Sciences, Beijing 100190, China and College of Information and Electrical Engineering, China Agricultural University,  Beijing 100083, China.}

\thanks{P. J. Yin is with the Institute of Applied Mathematics, Academy of Mathematics and Systems Science, Chinese Academy of Sciences, Beijing 100190, China.}

\thanks{Y. L. Li is with the State Key Lab of Management
	and Control for Complex Systems, Institute of Automation, Chinese Academy of Sciences, Beijing 100190, China}

\thanks{This work was supported in part by the National Natural Science Foundation of China under Grant 61210009.}}

\maketitle
\begin{abstract}
In recent years, the interdisciplinary research between information science and neuroscience has been a hotspot. Many biologically inspired visual and motor computational models have been proposed for visual recognition tasks and visuomotor coordination tasks.

In this paper, based on recent biological findings, we proposed a new model to mimic visual information processing, motor planning and control in central and peripheral nervous systems of human. Main steps of the model are as follows:

\begin{enumerate}
	
\item
\emph{Simulating "where" pathway in human}: the Selective Search method is applied to simulate the function of human dorsal visual pathway to localize object candidates; 

\item
\emph{Simulating "what" pathway in human}: a Convolutional Deep Belief Network is applied to simulate the hierarchical structure and function of human ventral visual pathway for object recognition;

\item
\emph{Simulating motor planning process in human}: habitual motion planning process in human is simulated, and motor commands are generated from the combination of control signals from past experiences;

\item
\emph{Simulating precise movement control in human}:  
calibrated control signals, which mimic the adjustment for movement from cerebellum in human, are generated and updated from calibration of movement errors in past experiences, and sent to the movement model to achieve high precision. 

\end{enumerate}

The proposed framework mimics structures and functions of human recognition, visuomotor coordination and precise motor control. Experiments on object localization, recognition and movement control demonstrate that the new proposed model can not only accomplish visuomotor coordination tasks, but also achieve high precision movement with learning ability. Meanwhile, the results also prove the validity of the introduced mechanisms. Furthermore, the proposed model could be generalized and applied to other systems, such as mechanical and electrical systems in robotics, to achieve fast response, high-precision movement with learning ability.
\end{abstract}


%
\IEEEpeerreviewmaketitle

\section{Introduction}
%
%
%
%
\IEEEPARstart{R}{obotics} research has made a lot of progress in recent years. Many different types of robots have been designed and developed, especially those with biologically inspired or human-like mechanisms and functions. For example, ECCE robot (Embodied Cognition in a Compliantly Engineered Robot) mimics the structures of human, which include bones, joints, muscles and tendons \cite{Marques2010}. ICub robot is designed to mimic a $3$ year old child, and has $53$ degrees of freedom (DOF) in total \cite{Metta2010}. With such complex structure of human-like robot, related biologically inspired computational models have also been developed, which mainly focused on vision, motor, and visuomotor coordination aspects.

In visual recognition tasks, many computational models have been proposed, which include Neocognitron model \cite{Fukushima1988}, saliency based visual attention model \cite{Itti1998,Itti2006}, HMAX model \cite{Cox2015, Tacchetti2015, Huang2011, Serre2007, Theriault2013, Qiao2014, Qiao2015}, deep learning neural networks  \cite{Huang2011,Yan2014,Lee2009,Krizhevsky2012,Girshick2014,Schroff2015,Qiao2016}, and etc. Among these models, HMAX model mimics ventral stream (from primary visual cortex to inferior temporal cortex) of visual cortex in primates, which has a feed-forward hierarchical structure. With alternation between convolution and max-pooling process, HMAX model could generate a set of position- and scale-invariant features for  later recognition. Recently, Deep Neural Networks (DNN) have also been widely applied for visual recognition. Due to its multi-layer structure and large training data sets, DNN exhibits good performance in various visual tasks.  In motion tasks, different biologically inspired models have been proposed based on findings in motor system of insects \cite{Palmer2015}, primates \cite{Koo2015,Srinivasan2006} and human \cite{Kwon2014,Hunt2015}. Most models  mimic one specific function or gait of the organism, such as climbing \cite{Palmer2015}, walking \cite{Koo2015,Renjewski2015}, running \cite{Srinivasan2006} and etc. But the used mechanisms in these models are quite different from those in organisms. Moreover, it might limit the  compatibility of the model to be applied in other tasks. Thus, inner structure of the motor system (such as spindle, muscle, spinal cord and etc.) should be considered for a more bionic model to mimic movements of the animals. Recently, some progress has been made in this direction, such as human upper extremity model with proper muscle configuration \cite{Qiao20162}. In visuomotor coordination tasks, which are mostly visually-guided reaching or grasping tasks, biologically inspired models are proposed for the learning process, motor-primed visual attention, movement control with visual feedback signals and etc \cite{Horaud1998,Law2014,Lukic2015,Lopes2005}.

In this paper, based on recent biological findings, we propose a new model for object localization, recognition, motion planning, and movement calibration task, which mimics the mechanisms and functions in human central and peripheral systems. Here, grasping a badminton with four steps is taken as an example to evaluate the performance of the proposed model. The framework of the model mainly includes two processes.
\subsubsection{\textbf{Vision process}}
\emph{Mimicking two visual pathways in human, object localization and recognition are processed in two distinct ways}. 

In object localization, selective search method is applied and a classifier is trained to select proper bounding boxes for all object candidates. In object recognition, an unsupervised DNN model is applied to extract key features of the object, which can be shown by visualization of connection weights in the network. A classifier is then trained for object recognition from these extracted key features.

\subsubsection{\textbf{Motion process}}
\emph{In this process, motion planning and movement calibration are carried out in sequence}.

Mimicking human habitual planning theory, control signals in motion planning are not directly calculated via inverse dynamics, but estimated by linear combination of control signals from past experiences. Movement calibration, which mimics the main function of cerebellum in human, is achieved by learning from past movement errors and calculating 
corrected signals for the new movement target with high precision.

The rest of this paper is organized as follows. In section II, related biological evidence is reviewed and discussed. In section III, the framework and detailed description of the new model are presented. In section IV, the performance of the model is evaluated on badminton-grasping task, and the results are analyzed. In section V, conclusions are drawn and possible future research directions are discussed.

\section{Biological Evidence}
Since the proposed framework aims at mimicking information processing in human, related biological evidences are reviewed in this section, which mainly focus on visual processing, motion planning and precise movement control.

\subsection{Two distinct visual pathways in human brain}
In primate visual system, two types of information ("what" and "where") is processed in two distinct but interactive pathways: ventral and dorsal pathway \cite{Goodale1992}. In anatomy, the ventral pathway consists of $V1$, $V2$, $V4$, PIT (posterior infero temporal) and AIT (anterior infero temporal) area in the brain. Visual information enters the ventral pathway from primary visual cortex and transfer along the rest areas in sequence. The main function of ventral pathway is highly associated with object recognition \cite{Lamme1998,Tanaka1993}. Meanwhile, the dorsal pathway also starts from primary visual cortex, but continues in V2, V3, MT (middle temporal), MST (medial superior temporal), LIP (lateral intraparietal sulcus) and VIP (ventral intraparietal sulcus) area. The dorsal pathway is involved in spatial awareness and guidance of actions \cite{Bear2007}.
In function, ventral stream provides abstract representations of the environment, stores related information for later references, and helps to plan actions "off-line". While dorsal stream responds in real time, which could guide the programming of related actions at the instant. In summary, two pathways of visual information processing is designed for perception and action, respectively \cite{deHaan2011,Ungerleider1994}.

\subsection{Motion planning in human motor cortex}
In primates, primary motor cortex (M1) plays an important role in movement planning \cite{Sanes1995, Blohm2009}. Several experiments on $M1$ in monkeys have proved that when monkeys make an arm movement to reach for a target, the neurons in $M1$ are tuned to the direction of movement \cite{Georgopoulos1986,Arce2010}. In the process, each neuron showed maximal firing activity when the movement direction is its preferred one. Thus, a population vector could be constructed from firing activities of many neurons in M1 to predict the hand movement direction. It provides evidence for population coding strategy in the movement system.

\subsection{Precise control of movement in human}	
In neurobiology, the main function of cerebellum is to ensure coordination and precision of the movement. Most related findings are from examination of patients without cerebellum in clinic. These patients are able to make movements, but the movement is acted in an unstable, uncoordinated way. Thus, the basic function of cerebellum has been exhibited as calibrating the detailed form of a movement \cite{Strick2009,Buckner2013,Therrien2015}.
Besides it basic function on precise control of a movement, cerebellum also contribute a lot to several types of motor learning, especially when it is necessary to make elegant adjustments to how an action should perform \cite{Boyden2004,Brooks2015}.

\section{Model structure and algorithms}
Based on the above mentioned biological evidences, the framework of the proposed model is illustrated in Fig. 1. The model consists of four steps: localization of object candidates, object recognition, motion planning and movement calibration. The first two parts on vision process are simulated in Matlab, while the last two parts on motion process are implemented in OpenSim platform, which include models of musculoskeletal structures and  simulation of dynamics during movements. In this section, function and detailed description of each block are presented.

\begin{figure*}[!htb]
	\centering
	\includegraphics[width=1.0\linewidth]{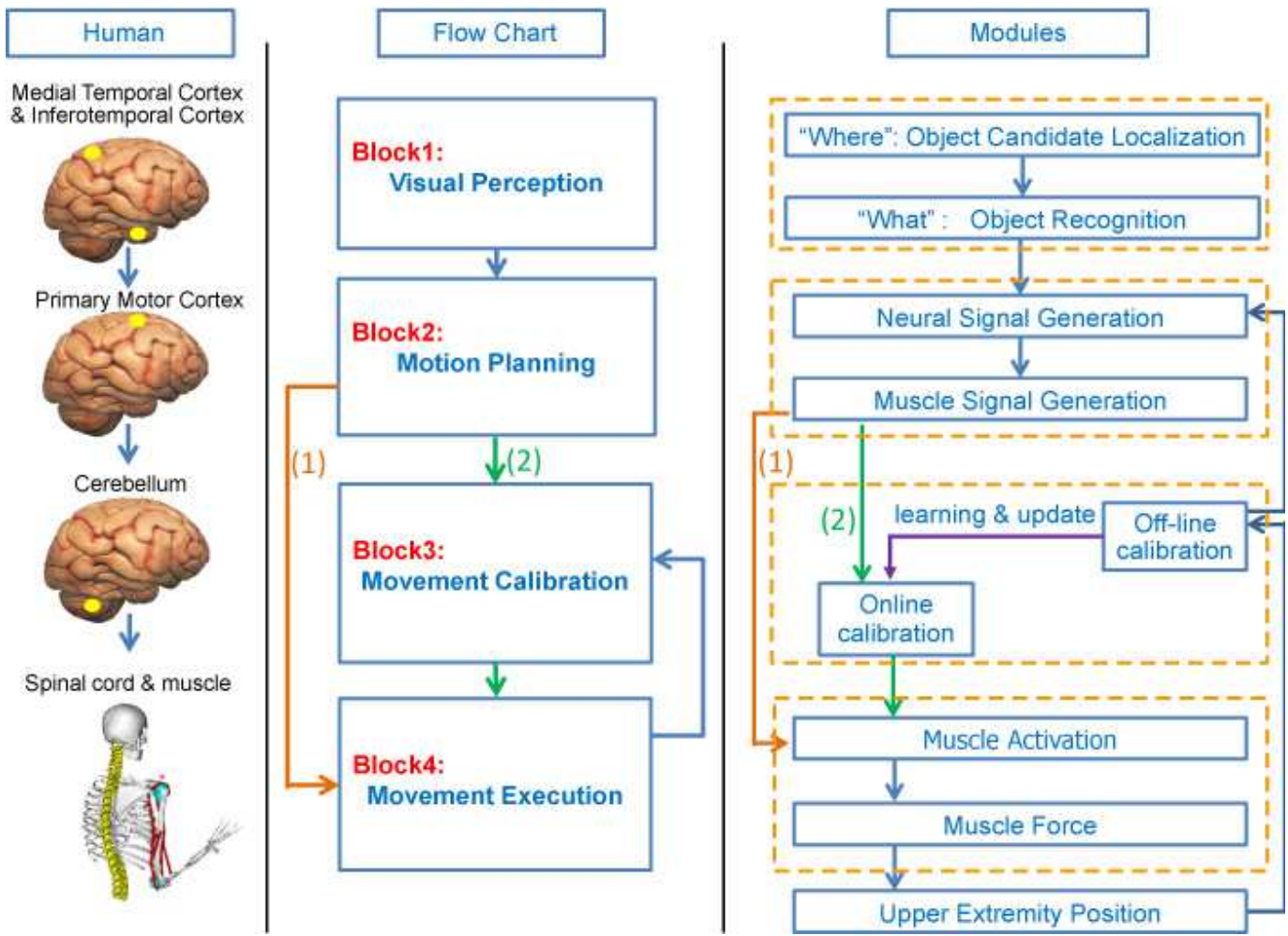}
	\caption{The Framework of the proposed model. In the left column, activation of related brain areas are shown for the task. In the middle column, corresponding flow chart is shown. On the right side, detailed modules in the model are presented.}
	\label{fig:framework}
\end{figure*}

\subsection{Block1: Visual perception -- "where" and "what"}
As reviewed in Section II, two distinct visual pathways contribute to different functions in human brain. Simulating properties of each pathway, the model for visual perception is also divided into two parts and described below.

\subsubsection{Block1-1: "Where" -- localization of object candidate}
In Block $1$, mimicking the function of dorsal pathway in human visual cortex, the positions of objects are achieved in two steps: 
bottom-up saliency extraction of object candidates, top-down segmentation of the region of interest. Based on biological findings, it is proposed that the activation of dorsal pathway is faster than that of ventral pathway \cite{Bullier2001,Klistorner1997}, which guarantees a fast response for movement without recognition. 

Bottom-up excitation comes from the stimulus. In other words, visual features (such as the color or shape of the object) at each location in the visual field could evoke strong responses of the neurons, and the integration of these features is formed for the possible locations of objects \cite{Treue2003}. Top-down modulation comes from higher hierarchical visual layer (such as ventral intraparietal areas), which suppresses the activity of neuron populations for non-attended attributes\cite{Riddoch2010}. 

Firstly, the selective search method is used for unsupervised extraction of object candidates \cite{Uijlings2013}. This method could generate a series of object proposals by integrating a variety of color space, texture and size features in a bottom-up hierarchical grouping of image segments, which corresponds to the hierarchical various feature encoding ability of visual cortex\cite{Treue2003}.

Secondly, although the selective search method can reduce the number of object candidates sharply than sliding window method, the number is still large. Here, a classifier is trained to further select the region of interests (RoIs). Positive training set comprises RoIs that have intersection union (IoU) overlap with a ground-truth general object bounding box of at least $0.8$, and the negative training set is sampled from the RoIs that have a maximum IoU with ground truth in the interval $[0, 0.5]$. Raw RGB pixels are taken as features. Finally, after a non-maximum suppression of the RoIs with high scores, the method outputs a few bounding boxes of object candidates, which guarantees a fast general object location and speeds up the computation for object recognition.

\subsubsection{Block 1-2: "What" -- object recognition}
In Block $2$, the function of the ventral pathway in visual cortex for object recognition is mimicked. The visual cortex in human is composed of many structurally and functionally different layers with many cortical-cortical connections, which form a hierarchical complex network \cite{Yang1992}. Moreover, object recognition is achieved in an unsupervised way, which suggests that temporal contiguity of object during natural visual experience can instruct the learning of the object features automatically \cite{Li2010,Stryker1991}.

A Convolutional Deep Belief Network (CDBN) organizes in a hierarchical structure is applied for unsupervised feature learning of object, as shown in Fig. 2. The CDBN model includes a visible layer ($\mathbf{V}$) and two convolutional restricted Boltzmann machines (CRBM) successively. As illustrated in \cite{Qiao2016}, the visualizations of the convolutional weights of the first and second CDBN correspond to edge detectors and key components, respectively. Biological findings also indicate that the V1 of visual cortex can discriminate small changes in visual orientations, and IT layer of visual cortex is tuned to components of object.

For details, one CRBM model consists of visible layer ($\mathbf{V}$), hidden layer ($\mathbf{H}$) and pooling layer ($\mathbf{P}$). $\mathbf{V}$ is the input of pre-processed image, and $\mathbf{H}$ and $\mathbf{P}$ both have $K$ groups of feature maps $\mathbf{H}^k (k= 1,2,\cdots,K)$ and $\mathbf{P}^k (k= 1,2,\cdots,K)$. The hidden layer $\mathbf{H}$ is connected with visible layer $\mathbf{V}$ in a local and weight sharing way. The structure of one CRBM model with the $\mathbf{k}$th channel is given in Fig. 2.

To simplify, we suppose the input image is square. The widths of the $\mathbf{V}$, the convolutional filter $\mathbf{W}$ and $\mathbf{H}$ are $n_v$, $n_w$, and $n_h$, respectively. By setting the convolutional step as $1$, $n_h$ equals to $n_v-n_w +1$. The width of $\mathbf{P}$ is $n_p= n_h /c$, and $c$ is the width of a pooling block. Thus $p^k_\alpha$ is obtained by pooling from a specific $c \times c$ block, denoted by $\alpha$. $v_{i,j}$ is a unit in $\mathbf{V}$, and $h_{i,j}^k$ is a unit in the $k$th feature map of $\mathbf{H}$, $i$ and $j$ corresponds to the row and column number in one feature map, respectively.

In all the experiments, the parameters of CRBMs are selected as $K=10$ and $c =2$, and the width of $\mathbf{W}$ is varied to verify whether different local features will affect the recognition performance.

Mathematically, the CRBM is a special type of energy based models \cite{LeCun2006}. When dealing with real inputs and binary hidden feature maps, the energy of each possible state ($v$, $h$), where $v\in \mathbb{R}^{n_v \times n_v}$ and $h\in \mathbb{B}^{n_h \times n_h \times K}~(\mathbb{B}=\{0,1\})$, is defined as:

\begin{eqnarray}
E(v,h) &=& -\sum_{k=1}^{K}\sum_{i,j=1}^{n_h}h_{i,j}^k (\mathbf{\tilde{W}}^k * v)_{i,j} -\sum_{k=1}^{K}b_k \sum_{i,j=1}^{n_h} h_{i,j}^k  \nonumber  \\
& &   - a\sum_{i,j=1}^{n_v}v_{i,j} + \frac{1}{2}\sum_{i,j=1}^{n_v} v_{i,j}^2,  \label{eq:energy}
\end{eqnarray}

where $h_{i,j}^k$ satisfies the constraint

\begin{equation*}
\sum_{(i,j)\in B_{\alpha}} h_{i,j}^k \leq 1, \forall k,\alpha. \hspace{1.18in}
\end{equation*}

Here, $\mathbf{\tilde{W}}^k$ denoting the $180$-degree rotation of the convolutional weights $\mathbf{W}^k$, * denotes the convolution operation, $b_k$ is the shared basis of all units in $\mathbf{H}^k (k=1,2,\cdots,K)$, and $a$ is the shared basis of visible layer units. The constraint condition will be used in the inference procedure of the CRBM.

Here, the two CRBMs are trained with Contrastive Divergence (CD) and approximate maximum-likelihood learning algorithm \cite{Hinton2002} in sequence. More details can be found in \cite{Qiao2016}.

After the training of the CDBN, the mean values over the $K$ feature maps of the $\mathbf{P2}$ layer belong to the second CRBM are computed as (\ref{mean}). The new feature map is named as $\mathbf{MP2}$ because of the mean operation.

\begin{equation}
\mathbf{MP2}_{i,j} = \frac{1}{K}\sum_{k=1}^{K} \mathbf{P2}^{(k)}_{i,j} \label{mean}
\end{equation}

The $\mathbf{MP2}$ layer is taken as an efficient feature of input image, and used to train a classifier to achieve object classification. In the test process, each test sample will get a probability score that tells its chance to be badminton.
\begin{figure}[!htb]
	\centering
	\includegraphics[width=1.0\linewidth]{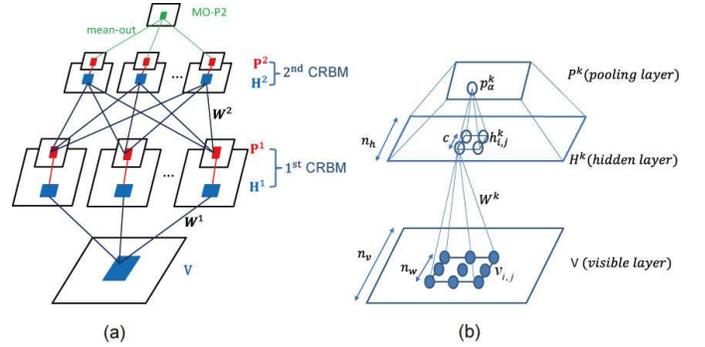}
	\caption{The network model for object recognition is illustrated. (a) Structure of the Deep Neural Network model is shown, which consists of a Convolutional Deep Belief Network (CDBN) and a mean-out pooling layer. The CDBN consists of two convolutional restricted Boltzmann machines (CRBM). The blue lines stand for the convolution, the red lines represent the probabilistic max-pooling, and the green lines represent the mean-out operation.(b) Structure of one CRBM with probabilistic max-pooling is shown. For simplicity, only the $kth$ channels of layer $\mathbf{H}$ and $\mathbf{P}$ are shown. Best view in electronic format.}
	\label{fig:CDBN}
\end{figure}

\subsection{Block 2: Planning of corresponding movement}  
In Block $3$, mimicking motion planning process in human, the habitual planning theory is applied in the model. In biology, two hypotheses on human movement planning are proposed: optimal control theory \cite{Diedrichsen2010} and habitual planning theory \cite{DeRugy2012}. Optimal control can minimize costs with respect to the effort, but it is rarely observed in human movement system. The habitual planning theory is proposed based on the fact that human tends to use past experience for the control of muscle contraction for the new movement. It could save computation from avoiding inverse kinematics calculation, and achieve rapid response.
 
Meanwhile, based on the study on primary motor cortex of monkeys, it is proposed that firing activities of a group of neurons can predict the movement direction of the arm \cite{Georgopoulos1986}. Hence, in order for the hand to access a new target, the excitation signals of the muscles in the upper extremity can be calculated based on those for previous training samples \cite{Qiao20162}. In this paper, previous training samples are defined as templates. Hence, the excitation signals of muscles for the movement of the new target should be calculated as:
\begin{equation}
 u_t(t) = \sum_{i=1}^N w_i u_{i}(t),
\end{equation}
where $u_t(t)$ is the excitation signals of muscles for the position of the target, $u_{i}(t)$ is the excitation signals of muscles for past movement for position $\bm{p_i}$, $w_i$ is the weight representing the contribution from each template to the target. In convenience, the motion planning is expressed in terms of the excitation signals of the corresponding muscles. 
 
In our previous work \cite{Qiao20162}, it is proved that the movement of the arm is continuous within a small area, which implies that two similar excitation signals of muscles lead to nearby positions. Hence, the weight $w_i$ in equation $3$ could be used to express the position of the target in terms of positions of the templates as follows:
\begin{equation}
 \bm{p_{t}} = \sum_{i=1}^N w_i \bm{p_i},
\end{equation}
 
 where $\bm{p_{t}} \in \mathbb{R}^3$ is the position of the target, $\bm{p_i} \in \mathbb{R}^3$ is the position of the template, $N$ stands for number of templates used for estimation of target, $w_i$ is  approximately calculated as:
\begin{equation}
 w_i \approx \frac{\frac{1}{d_{ti}}}{\sum_{j=1}^{N}\frac{1}{d_{tj}}},
\end{equation}
where $d_{ti}$ and $d_{tj}$ stand for the $L2$ norm of the vector  $\bm{p_t}-\bm{p_i}$ and $\bm{p_t}-\bm{p_j}$, respectively. The $L2$ norm is defined as $d_{ti}= ||\bm{p_t}-\bm{p_i}||_2$. 
 
\begin{figure}[!htb]
	\centering
	\includegraphics[width=1.0\linewidth]{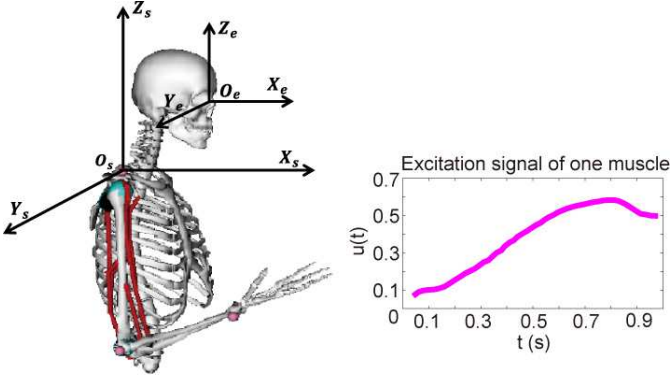}
	\caption{Motion planning of human upper extremity model. On the left side shows two coordinate systems in the model, which is centered with shoulder and the other with the eyes. On the right side illustrate the excitation signal of one muscle in the model during the movement.}
	\label{fig:planning}
\end{figure}

\subsection{Block 3: Precise movement control} 	
As previously mentioned in Section II, the main function of the cerebellum in human is to achieve movement with precision. 

Two types of calibration (off-line and online) are required in the cerebellum for the movement with high precision. Since the motor cortex in the brain sends out abstract signals  to the cerebellum for coarse movement, learning in the cerebellum is required to ensure the precision for different movements. Experiments on humans have shown that motor learning with cerebellum requires trial-and-error practice. When the behavior becomes adapted as learned, it is performed automatically \cite{Kandel}. Hence, the "trial-and-error practice" is the off-line calibration, while the "automatically-adapted behavior" is known as the online calibration. Moreover, the transition from off-line to online is based on the learning process, which establishes and updates the online calibration based on past experience of off-line calibration.

In neuroscience, specific neural circuits provide biological basis for the off-line and online calibration. The output projections of the cerebellum are mainly (a) on the premotor and motor area of the brain, and (b) on the brain stem to control spinal cord for the movement. Off-line calibration is then proposed to take place in (a), which is the "brain-cerebellum-brain" circuit \cite{Passot2010,Mitsunari2011}; while online calibration is considered as the function of projection (b), which is the "brain-cerebellum-spinal cord" circuit \cite{Cantarero2015,Thach1996}. The detailed description of off-line and online calibration is shown below.

\subsubsection{Off-line calibration: error correction for each movement}
According to Block $3$ of this section, motor commands are generated as the combination of weighted motor signals of the used templates. Since the motion model is a highly non-linear and coupled system, which is described in details in Block $5$, the combination of the  excitation signals of muscles cannot achieve the precise target position \cite{Qiao20162}. Hence, the error of the movement should be corrected to achieve movement with high precision. 

Since the new motor learning with cerebellum in human requires trial-and-error practice, which fits the off-line calibration regime, the error $\bm{e} \in \mathbb{R}^3$ of the movement should be corrected based on the target position $\bm{p_t}$ and actual position $\bm{p_a}$ of the movement.

It is proposed that the movement direction of human hand can be predicted by the firing activities of groups of neurons in motor cortex \cite{Georgopoulos1986}. The contribution of each individual neuron to the movement is represented as a vector along its preferred direction. Thus, the sum of these vectors can predict the movement direction of the hand, which is known as population vector coding \cite{Kazuyoushi2004,Jean2007}.

According to the above mentioned biological mechanisms, the error of the movement could be considered as improper contributions of individual neurons. Thus, the movement could be calibrated by adjusting the contributions of the used templates in the model. Based on the idea of population vector coding, off-line  calibration is to decompose the error of the movement into the weights of the excitation signals of used templates. The correction of the weights on each template $\Delta w_i$ is designed as
\begin{equation}
\Delta w_i^{off}=k_i \cdot \cos(\theta_{i}),
\end{equation}
where $\theta_{i}$ represents the angle between error vector $\bm{e}$ and the vector  $\bm{r_i} = \bm{p_i}-\bm{p_a}$, $k_i$ is the coefficient for each template and can be expressed as
\begin{equation}
k_i = \dfrac{d_{e}}{d_{a}} \cdot \left[ 1+n \cdot \left( \dfrac{d_t-d_{i}}{d_{i}} \right) \right] ,
\label{error1}
\end{equation}
where $d_e$ denotes $L2$ norm of error $\bm{e}$, $d_a$ and $d_t$ represents $L2$ norm of actual position $\bm{p_a}$ and target position respectively, ${d_i}$ represents the norm of the template $\bm{p_i} $, $n$ is a coefficient that is selected within $[0,20]$ to minimize $d_e$.

The corrected weight for each template is defined as $\mathop{w_i}'$: $\mathop{w_i^{off}}=w_{i}+\Delta w_i^{off}$, and the new excitation signal for each muscle  is then defined as $\mathop{u_{t}}'(t)$: $\mathop{u_{t}}'(t)=\sum_{i=1}^{N} \mathop{w_i^{off}} \cdot u_{i}(t)$, where $u_i(t)$ is the excitation signal of the muscle for template $i$.

\begin{figure}[!htb]
	\centering
	\includegraphics[width=1.0\linewidth]{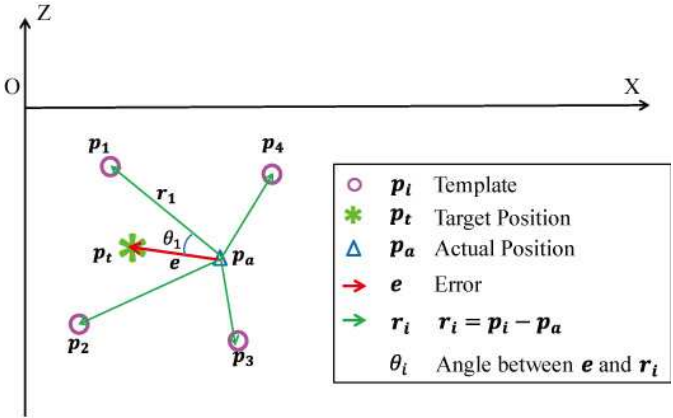}
	\caption{Illustration of the off-line calibration of the model. The error of the movement $\bm{e}$  could be decomposed into the contribution of each template. Details can be found in the context.}
	\label{fig:calibration}
\end{figure}

\begin{algorithm}[ht]	
	\caption{Off-line calibration of movement errors}
	\SetKwInOut{IN}{Input}
	\SetKwInOut{OUT}{Output}
	\IN{Target position $\bm{p_t}\in \mathbb{R}^3$, actual position of the movement $\bm{p_a}\in \mathbb{R}^3$, positions of the templates $\bm{p_i}\in \mathbb{R}^3$ and  corresponding weights $w_i$ $(i= 1,2,3,...,N)$}
	\OUT{Off-line calibrated weights $\mathop{w_i}^{off}$}
	\scalebox{0.96}{\parbox{\linewidth}{
			\begin{algorithmic}[1]
			    \STATE Define vector $\bm{r_i}=\bm{p_i}-\bm{p_a}$ and angle $\theta _i$ between $\bm{e}$ and $\bm{r_i}$
				\FOR {$n=1,2,3,...,M$} 
				\STATE Calculate the coefficient $k_i^n$ with Eq. ($7$)				
				\STATE Estimate calibrated value of weights $\Delta w_i^{n}$ with Eq. ($6$)			
				\STATE Apply forward dynamics with modified weights $\mathop{w_i}^{n}= w_i+\Delta w_i^{n}$ and calculate movement error $e^{n}$
				\ENDFOR							
				\STATE Select the minimal $e^{n}$ and the corresponding weights $\mathop{w_i}^n$ as the off-line calibration weights $\mathop{w_i}^{off}$				
			\end{algorithmic}
		}}
		\label{Alg1}
\end{algorithm}

\subsubsection{Online calibration: automatic correction for the new movement}
After the trial-and-error practice, the cerebellum could generate online adjusted signals to the spinal cord to achieve movement with high precision. This implies that cerebellum could learn the general relationship between the position of the target, motor commands and past experiences on error correction.  Mimicking this learning ability of the cerebellum, a general online calibration model for the automatic correction for the new movement is proposed based on the results of past off-line calibration, which is expressed as
\begin{equation}
\Delta w_i^{on} = g_i(\bm{p_t},w_1^{off},\bm{p_1^{off}},...,w_N^{off},\bm{p_N^{off}}),
\end{equation}

where $\bm{p_t} \in \mathbb{R}^3$ stands for the position of the target, $\bm{p_1^{off}}$, $\bm{p_2^{off}}$,...,$\bm{p_N^{off}} \in \mathbb{R}^3$ represent the positions of the movement after off-line calibration, $w_1^{off}$, $w_2^{off}$,...,$w_N^{off}$ represent the weight for each template after off-line calibration, $g_i$ is the linear regression model, which is built to estimate the online adjusted weights from target position, positions of the templates, and weights of the templates. Thus, the corrected weight of each template could be achieved via equation $(8)$ to ensure the new movement with high precision. Furthermore, the online calibration model $g$ is updated with the increasing number of movements.

\begin{algorithm}[ht]	
	\caption{Updating online calibration model}
	\SetKwInOut{IN}{Input}
	\SetKwInOut{OUT}{Output}
	\IN{Target position $\bm{p_t}\in \mathbb{R}^3$, positions of movements with off-line calibration $\bm{p_i^{off}}\in \mathbb{R}^3$, and corresponding  weights $w_i^{off}$ $(i= 1,2,3,...,N)$}
	\OUT{Updated model of online calibration $g'$}
	\scalebox{0.96}{\parbox{\linewidth}{
			\begin{algorithmic}[1]
				\STATE Estimate adjusted weights $\Delta w_i^{on}$ for online calibration with Eq. ($8$)
				\STATE Apply forward dynamics with calibrated weights $\mathop{w_i}^{on}= w_i+\Delta w_i^{on}$ and calculate movement error $e^{on}$
				\STATE Calibrate movement error	$e^{on}$ with off-line calibration and get  modified weights $\Delta w_i^{off}$
				\STATE Apply forward dynamics with weights $\mathop{w_i^{'}}=\mathop{w_i}^{on}+\Delta w_i^{off}$ and get the actual position $\bm{p_a^{'}}$ of the movement
				\STATE Include the modified weights $\mathop{w_i^{'}}$ and position $\bm{p_a^{'}}$ to update the online calibration model to $g'$
			\end{algorithmic}
		}}
		\label{Alg1}
	\end{algorithm}

\subsection{Block 4: Movement model of the upper extremity}  
The simplified model of the upper extremity has two joints and six muscles. The muscles embedded in the model are: long and short head of the biceps (BIClong, BICshort), brachialis (BRA), and three head of the triceps (TRIlat, TRImed, TRIlong) \cite{Garner2003}. These are amongst the most important muscles involved in performing the movement of arm, and the two joints are the elbow and the shoulder \cite{Davis1999}. Thus, the movement of the arm could be modeled in three parts: activation dynamics, musculotendon contraction dynamics and the motion of the upper limb \cite{Thelen2003,Thelen2006}.

Activation dynamics is the process to simulate muscle-fiber calcium concentration, which is modulated by firing activities of motor units. It is modeled as:

\begin{equation}
\mathop {\rm{\dot a}}\left( t \right)  = \left\{ {\begin{array}{*{20}{c}}
	{\left( {u\left( t \right) - a\left( t \right)} \right)\left[ {\frac{{u\left( t \right)}}{{{\tau _{act}}}} + \frac{{1 - u\left( t \right)}}{{{\tau _{deact}}}}} \right],u\left( t \right) \ge a\left( t \right)}\\
	{({u\left( t \right) - a\left( t \right)})/{{\tau _{deact}}},u\left( t \right) < a\left( t \right)}
	\end{array}} \right.
\end{equation}	
where $u(t)$ is the excitation signal of the muscle in general, $a(t)$ is the activation of the muscle, ${\rm{\dot a}}\left( t \right)$ is the change rate of the activation of the muscle,  $\tau_{act}$ and $\tau_{deact}$ are the time constants for activation and deactivation, respectively.

Musculotendon contraction dynamics is the process to calculate the muscle forces and it is expressed as:

\begin{equation}
{\bm{F_m}} = {\bm{F_0}}\left( {{f_1}{f_2}a\left( t \right) + {f_3}} \right)
\end{equation}
where $\bm{F_m} \in \mathbb{R}^3$ represents the muscle force, $\bm{F_0} \in \mathbb{R}^3$ is the initial muscle force, $f_1$,$f_2$ and $f_3$ are sub-functions, which are calculated as:

\begin{equation}
\begin{array}{l}
{f_1} = {e^{\left[ { - 40{{\left( {l - 0.95} \right)}^4} + {{\left( {l - 0.95} \right)}^2}} \right]}}\\
{f_2} = 1.6 - 1.6{e^{\left[ {\frac{{ - 1.1}}{{{{\left( { - v + 1} \right)}^4}}} + \frac{{0.1}}{{{{\left( { - v + 1} \right)}^2}}}} \right]}}\\
{f_3} = 1.3{\rm{arctan}}\left[ {0.1\left( {{{\left( {l - 0.22} \right)}^{10}}} \right)} \right]\\
l = \frac{{{l_m}}}{{{l_0}}}\ and\ {\bm v} = \frac{{{\rm{}}{\bm{v_m}}}}{{2.5}}
\end{array}
\end{equation}
where $\bm{v_m} \in \mathbb{R}^3$ is the velocity of muscle contraction, $l_m$ is the length of muscle fiber, $l_0$ is the initial length of muscle fiber.

Motion of the model in response to the applied muscle forces is modeled as:
\begin{equation}
\mathop {\bm q}\limits^{..}  = {A^{ - 1}}\left( {\bm q} \right)\left\{ {R\left( {\bm q} \right){\bm {F_m}} + {\bm G}\left( {\bm q} \right)} \right\}
\end{equation}
where $\bm q \in \mathbb{R}^3$ is the generalized coordinates of the model, and $\mathop {\bm q}\limits^{..} \in \mathbb{R}^3$ is the accelerations. $A^{-1}$ is the inverse of system mass matrix, $\bm G \in \mathbb{R}^3$ is other environment forces, $R$ is a matrix of muscle moment arms.

To evaluate the motion in response to the neural excitation signals, the joint angles and the position of the hand can be calculated by integrating the equations above. In this paper, OpenSim is applied as the platform for the implementation of the model \cite{Pennestri2007}.

\section{Experiments and analysis}
To verify the biologically inspired model and algorithm proposed above, the task of detecting and grasping a badminton is taken as an example, which consists of four corresponding modules: localization of object candidates, object recognition, motion planning and movement calibration. The model is evaluated on CASIA-RTA-VM data set (established in our lab). In this section, the results of each module are presented and discussed.

\subsection{Localization of candidates of a badminton} 
The localization of candidates of a badminton is evaluated on CASIA-RTA-VM data set. The image in this database contains a badminton and a cup. Some samples of the database are shown in Fig. $5$. 

\begin{figure}[!htb]
	\centering
	\includegraphics[width=1.0\linewidth]{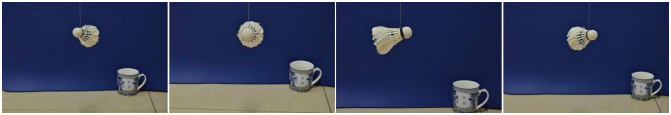}
	\caption{Illustration of images used in the experiments.}
	\label{fig:calibration}
\end{figure}

Firstly, selective search method is applied to $80$ images to extract the positions of candidates of badminton. The size of each image is $600 \times 400$ pixels. In average, $130$ bounding boxes are extracted via selective search method from each image. In total, $10560$ bounding boxes are selected for later processing. Since selective search combines the exhaustive search method and segmentation method \cite{Uijlings2013}, this algorithm is faster than exhaustive search by reducing the number of locations on the foundation of capturing all scales within the image. Meanwhile, selective search uses a diverse set of grouping strategies to make itself robust and independent of object-class.

Secondly, a classifier is trained to find the proper RoIs of the object with the LibSVM toolbox \cite{Chang2011}. Based on the results from last step, bounding boxes from $40$ images are selected for training, while those from the other $40$ images are chosen for testing. Ground truth is defined here as the properest bounding box that contains each candidate of the object in each image. In training data set, there are $224$ positive samples and $4707$ negative samples chosen according to the selection principles mentioned in Section III. In testing data set, $232$ positive samples and $4702$ negative samples are selected.
The experiments on classifiers with different kernels are conducted and the results are shown in Table I. Although only pixel features are used, classifier with linear kernel could achieve comparative results with Radial Basis Function kernel. The linear kernel is finally selected for later usage.
\begin{table}[!htb]
	\centering
	\caption{Test results of classifier with different kernels.}
	\label{tabel:kernel}
	\begin{tabular}{|c|c|c|}
		\hline
		Kernel function & Precision & Recall \\ \hline
		Gaussian kernel & 93.06\% & 86.64\% \\ \hline
    	Radial Basis Function kernel & 100\% & 93.53\% \\ \hline
		Polynomial kernel & 97.74\% & 93.10\% \\ \hline
		Linear kernel & 100\% & 93.53\% \\ \hline
	\end{tabular}
\end{table}

Thirdly, based on the classifier score of each selected RoI in previous step, a non-maximum suppression is applied. In other words, if the overlapping region between two bounding boxes is larger than $30 \%$, the one with a higher score are selected \cite{Neubeck2006}. Then   $80 \%$ is chosen as the threshold, which is the minimal percentage of the overlapped region between the selected RoI and ground truth. If the overlapped region is larger than the threshold, it is considered as a positive sample. The results are shown in Table II. The flow chart of localization of candidates is shown in Fig. $6$.

\begin{table}[!htb]
	\centering
	\caption{The recall and precision of our model on localization.}
	\label{tabel:localization}
	\begin{tabular}{|c|c|c|}
		\hline
		& Precision & Recall \\ \hline
		SS+classifier & 100\% & 94.94\% \\ \hline
	\end{tabular}
\end{table}

\begin{figure}[!htb]
	\centering
	\includegraphics[width=1.0\linewidth]{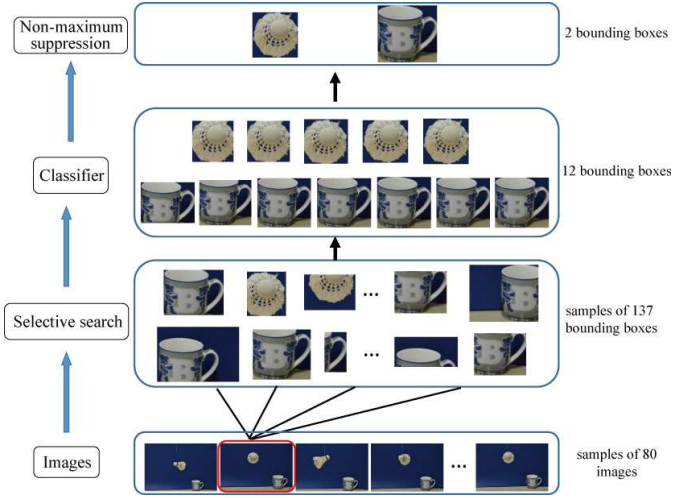}
	\caption{The framework of localization of object candidates. The examples of input image examples are shown on the bottom, and the image with red box is used for illustration of further processing steps. The number of the samples for each step is shown on the right side of the figure.}
	\label{fig:flowchart}
\end{figure}

\subsection{Recognition of a badminton} 
The CDBN model for unsupervised feature learning of the badminton is trained with the images within the selected bounding boxes from last step. Positive samples are chosen as the badminton, while the negative samples are chosen as the cup. In total, $80$ images of badminton and $80$ images of cups are used. The image size is resized to $200 \times 200$ pixels, and  $50,000$ patches with size $21 \times 21$ are randomly sampled in the preprocessing training set for the learning of the first CRBM of CDBN. Then $50,000$ patches with scale $25 \times 25 \times K1$ are randomly sampled in the $\mathbf{H}^{(1)}$ layer of the positive training set for the learning of the second CRBM. Thus, the CDBN model is trained and its visualization results are discussed below.

Firstly, the visualization results of the learned weights are illustrated in Fig. $7$. It can be observed from the result that $\mathbf{W^{1}}$ corresponds to the edges with different orientation preference, which is similar to Gabor filters of $V1$ layer of primate visual cortex \cite{Merigan1993}. The weight $\mathbf{W^{2}}$ corresponds to the special texture of the badminton, which indicates that the CDBN model learns more complex features step by step.

\begin{figure}[!htb]
	\centering
	\includegraphics[width=1.0\linewidth]{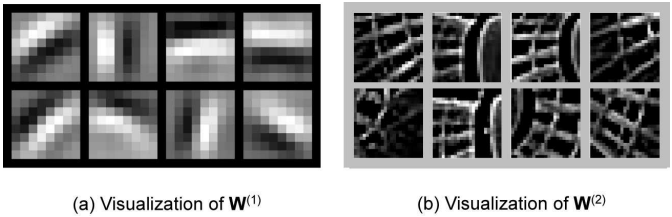}
	\caption{Visualization of the learned weights of the CDBN. The size of $\textbf{W}^{(1)}$ is 8@8$\times$8, and the corresponding size of $\textbf{W}^{(2)}$ in $\textbf{V}$ layer is 8@23$\times$23.}		
	\label{fig:calibration}
\end{figure}

Secondly, visualization of the feature maps of each layer in CDBN is shown in Fig. $8$. 
Since the pose of the badminton in each input image is not the same, the activation of feature maps is tuned to the specific textures of the badminton, which can be seen in $\mathbf{H}^{(1)}$ and $\mathbf{P}^{(1)}$.

\begin{figure}[!htb]
	\centering
	\includegraphics[width=1.0\linewidth]{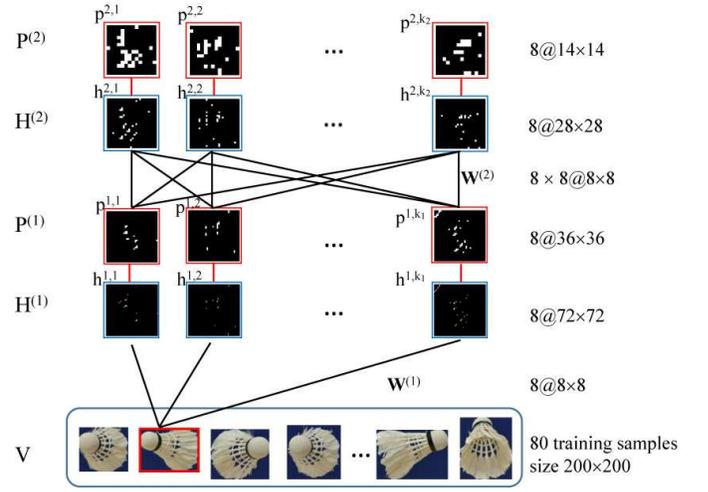}
	\caption{Visualization results of feature maps in CDBN. Input image examples are shown in $V$, and the image with red box is used for visualization of higher feature maps. The sizes of the feature maps are shown on the right side of the figure.}
	\label{fig:calibration}
\end{figure}

The new model is also compared with other methods on object recognition, such as SIFT \cite{Lowe1999} or HOG  \cite{Dalal2005} based models. Different from these hand-crafted features, CDBN model is an unsupervised feature learning model, which can learn discriminative features automatically. Comparing with HMAX model \cite{Serre2007}, which is also unsupervised, "filter templates" of the HMAX model limits its learning ability. The comparison experiment between CDBN and HMAX model is carried out. In HMAX model, $10$ 'filter templates' with scale $4$ are selected, which are similar as $n_w=8$ in CDBN model for different pooling scales. The results are given in Table III.

\begin{table}[!htb]
	\centering
	\caption{The recall and precision of CDBN model and the HMAX model.}
	\label{tabel:classification}
	\begin{tabular}{|c|c|c|}
		\hline
		& Precision & Recall \\ \hline
		CDBN & 98.8\% & 100\% \\ \hline
		HMAX & 97.5\% & 100\% \\ \hline
	\end{tabular}
\end{table}

\subsection{Motion planning for grasping a badminton} 
As described in Section III. C, motion planning is achieved based on past experience. The method is proposed in $3$D space. Here, the movement in 2D space is applied to evaluate its validity. The excitation signal of one muscle is computed as Eq. ($4$), which is shown in Fig. $9$. Here, the number of templates to evaluate the excitation signal of one muscle is chosen as $N = 4$.

\begin{figure}[!htb]
	\centering
	\includegraphics[width=1.0\linewidth]{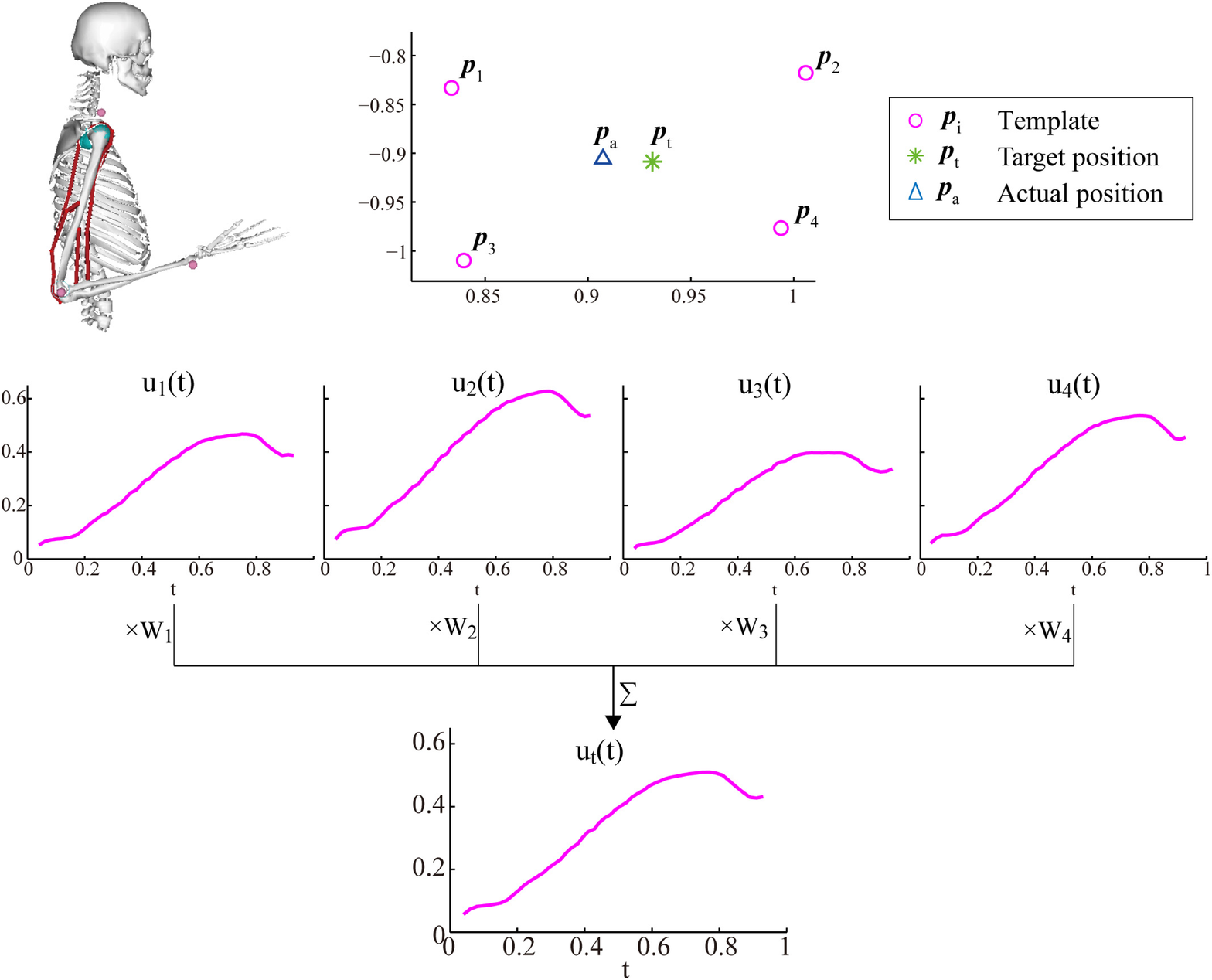}
	\caption{Motion planning of the target. The excitation signal of one muscle is taken as an example to calculate as the weighted sum of those of the templates. The actual position of the movement is achieved with all the excitation signals of muscles, which is not exactly the same as the position of the target.}
	\label{fig:calibration}
\end{figure}

With the calculated excitation signals of all the muscles, the movement of human upper extremity is achieved with forward dynamics of Eq. ($9$)-($12$) in OpenSim. Since the excitation signal of each muscle is not directly calculated from inverse kinematics, this method saves computation and leads to faster response. Meanwhile, with this approximate estimation of the excitation signals of the muscles, the actual position of the movement is not exactly the same as the position of the target, especially when the number of templates is small. The errors of the movement are shown in Table IV.

\begin{table*}[!htb]
	\centering
	\caption{Error of the movement.}
	\label{tabel:movement error}
	\begin{tabular}{|c|c|c|c|c|c|}
		\hline
		Target position & (0.8634,-0.7487) & (0.9132,-0.7849) & (0.8565,-0.7940) & (0.9518, -0.7336) & (0.8906,-0.7698)\\ \hline
		Actual position & (0.8594,-0.7336) & (0.8734,-0.7262) & (0.8499,-0.7588) & (0.8983, -0.6907) & (0.8581,-0.7230) \\ \hline
		\multirow{4}{*}{Calculated weights} & 0.2677 & 0.3172 & 0.1673 & 0.1848 & 0.3473 \\ \cline{2-6} 
		 & 0.3340 & 0.4503 & 0.3002 & 0.2853 & 0.2367 \\ \cline{2-6} 
		 & 0.2095 & 0.2167 & 0.2113 & 0.3680 & 0.2131 \\ \cline{2-6} 
		 & 0.1888 & 0.1351 & 0.1712 & 0.1619 & 0.2029 \\ \hline	 
		Movement error & 0.0156 & 0.0709 & 0.0357 & 0.0686 & 0.0569 \\ \hline
		Mean error & \multicolumn{5}{c|}{0.0496} \\ \hline	
		{\tabincell{c}{\textbf{Mean error} \\ (all samples)}} & \multicolumn{5}{c|}{0.0393} \\\hline		
	\end{tabular}
\end{table*}

\subsection{Precise control of grasping} 
As described in Section III. D, off-line calibration is achieved via "trail-and-error practice", which is carried out after the movement; while online calibration is based on past results of off-line calibration, which is achieved before the movement.

\subsubsection{Off-line calibration}
Based on the actual position of the movement in last step, the error of the movement is decomposed into weights of the excitation signals of the templates. The adjusted weights are calculated from Eq. ($6$)-($7$) and then the calibrated excitation signals are derived. Forward dynamics of the excitation signals are calculated and examples of the calibrated positions of the movements are shown in Fig. $10$. 

\begin{figure}[!htb]
	\centering
	\includegraphics[width=1.0\linewidth]{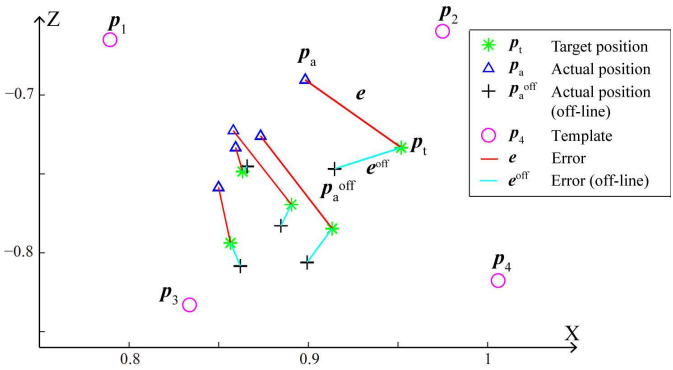}
	\caption{Examples of off-line  calibration. $5$ examples are shown in the figure, which correspond to Table V. For the same target, the error of calibrated movement is smaller.}
	\label{fig:calibration}
\end{figure}

The examples of calibrated weights and positions are given in Table V. The mean error based on $25$ samples of the movement is smaller after the calibration, which illustrates the effectiveness of off-line calibration.

\begin{table*}[!htb]
	\centering
	\caption{Off-line   calibration of the movement.}
	\label{tabel:Off-line correction}
	\begin{tabular}{|c|c|c|c|c|c|}
		\hline
		Target position & (0.8634,-0.7487) & (0.9132,-0.7849) & (0.8565,-0.7940) & (0.9518, -0.7336) & (0.8906,-0.7698)\\ \hline
		Actual position & (0.8659,-0.7455) & (0.8993,-0.8063) & (0.8621,-0.8086) & (0.9146, -0.7471) & (0.8846,-0.7833) \\ \hline
		\multirow{4}{*}{Calculated weights} & 0.2470 & 0.3589 & 0.1115 & 0.2018 & 0.3814 \\ \cline{2-6} 
		& 0.3435 & 0.3295 & 0.5630 & 0.3143 & 0.2534 \\ \cline{2-6} 
		& 0.2061 & 0.2125 & 0.1221 & 0.4019 & 0.2185 \\ \cline{2-6} 
		& 0.1905 & 0.0438 & 0.1659 & 0.0444 & 0.1058 \\ \hline	 
		Movement error & 0.0040 & 0.0256 & 0.0157 & 0.0395 & 0.0147 \\ \hline
		Mean error & \multicolumn{5}{c|}{0.0199} \\ \hline		
	 	{\tabincell{c}{\textbf{Mean error} \\ (all samples)}} & \multicolumn{5}{c|}{0.0234} \\\hline				
	\end{tabular}
\end{table*}

\subsubsection{Online calibration and learning}
Based on the results of off-line calibration, online calibration is achieved with Eq. ($8$). The adjusted weights and positions are shown in Table VI. Examples of the calibrated positions of the movements are shown in Fig. $11$. The mean error based on $25$ samples of the movement is smaller after online calibration, and it is similar with that of off-line calibration, which proves the validity of the online calibration model.

\begin{figure}[!htb]
	\centering
	\includegraphics[width=1.0\linewidth]{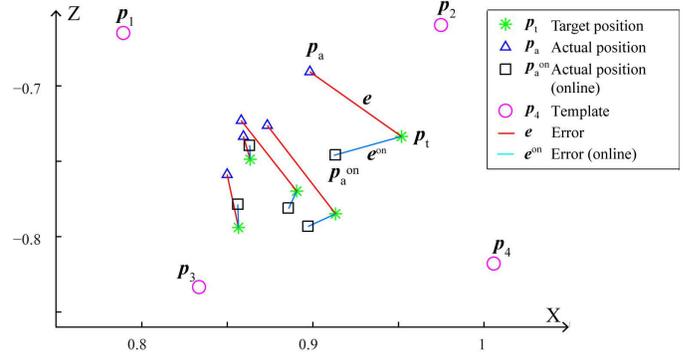}
	\caption{Examples of online calibration. $5$ examples are shown in the figure, which correspond to Table VI. }
	\label{fig:calibration}
\end{figure}

\begin{table*}[!htb]
	\centering
	\caption{On-line  calibration of the movement.}
	\label{tabel:On-line correction}
	\begin{tabular}{|c|c|c|c|c|c|}
		\hline
		Target position & (0.8634,-0.7487) & (0.9132,-0.7849) & (0.8565,-0.7940) & (0.9518, -0.7336) & (0.8906,-0.7698)\\ \hline
		Actual position & (0.8631,-0.7394) & (0.8972,-0.7933) & (0.8561,-0.7787) & (0.9123,-0.7461) & (0.8858, -0.7814)  \\ \hline
		\multirow{4}{*}{Calculated weights} & 0.2666 & 0.3166 & 0.1575 & 0.1794 & 0.3545 \\ \cline{2-6} 
		& 0.3365 & 0.3255 & 0.5504 & 0.3179 & 0.2599 \\ \cline{2-6} 
		& 0.2078 & 0.2210 & 0.1293 & 0.3884 & 0.2241 \\ \cline{2-6} 
		& 0.1795 & 0.0851 & 0.1482 & 0.0749 & 0.1186 \\ \hline	 
		Movement error & 0.0093 & 0.0181 & 0.0152 & 0.0414 & 0.0125 \\ \hline
		Mean error & \multicolumn{5}{c|}{0.0193} \\ \hline		
		{\tabincell{c}{\textbf{Mean error} \\ (all samples)}} & \multicolumn{5}{c|}{0.0238} \\\hline				
	\end{tabular}
\end{table*}

Furthermore, the online calibration model is updated based on the results of past off-line and online calibration. Results of the updating of the online calibration is shown in Fig. $12$. Each experiment is composed of $25$ movement tasks. In the first experiment, the online calibration model is built based on the results of past off-line calibration. After the online-calibrated movement, the error still exists and it is calibrated with off-line calibration method. With these calibrated weights $w_i^{'}$ of $25$ movements, the online calibration model is updated with Eq. ($8$). From Fig. $12$, it is clear that the movement error decreases with increasing number of experiments.

\begin{figure*}[!htb]
	\centering
	\includegraphics[width=\linewidth]{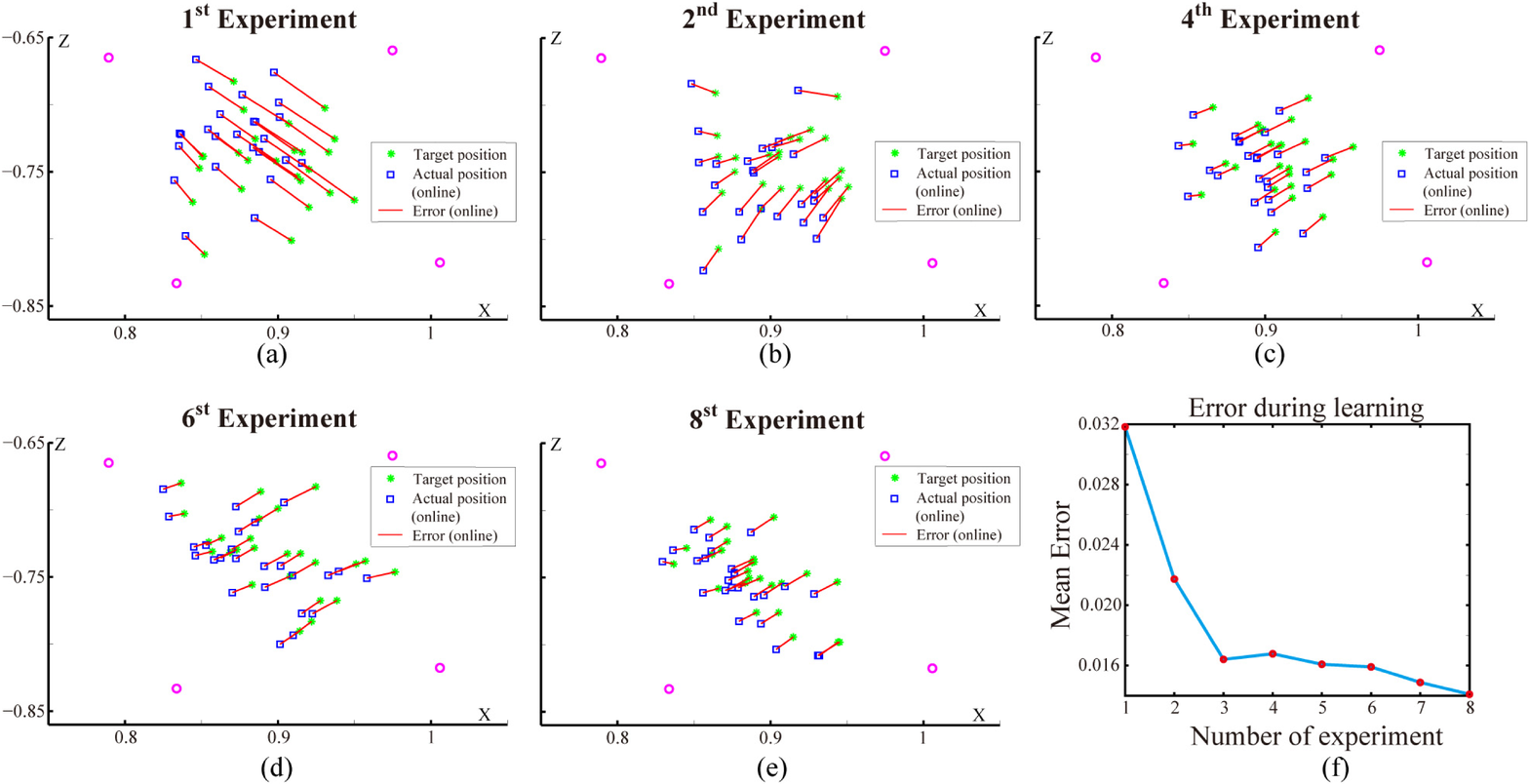}
	\caption{Update of the online calibration model. (a)-(e) Examples of movement with updated online calibration model. (f) Mean error of the movement decreases with update of online calibration model.}
	\label{fig:neuro-robot}
\end{figure*}

\section{Conclusion}
In this paper, a new visuomotor coordination model based on related mechanisms in human visual processing, motor planning and precise control is proposed. The model exhibits its abilities on visuomotor coordination, off-line and online calibration of the movement, which can accomplish motion tasks in a precise way with learning ability. 

The proposed model has four main functions: localization of object candidates, object recognition, motion planning and movement calibration. The localization of object candidates and object recognition are achieved with two distinct methods, which simulate two visual pathways in human. Motion planning applies human habitual movement planning theory, while movement calibration mimics the function of cerebellum in human. This visuomotor-integrated model could achieve fast perception and response, off-line and online adjustment of movement, and learning ability of precise motor control. Especially, the learning ability plays a crucial role in the updating of online calibration.

Furthermore, the proposed model provides a general framework of visuomotor coordination and precise control for complex systems, which could be extended and applied to robotic system to verify its validity and performance. In our lab, a neuro-robot, which mimics human movement system with muscle-tendon structure, is designed and built up. In the future, the proposed model will be implemented to this system to test its efficiency.

\begin{figure}[!htb]
	\centering
	\includegraphics[width=0.7\linewidth]{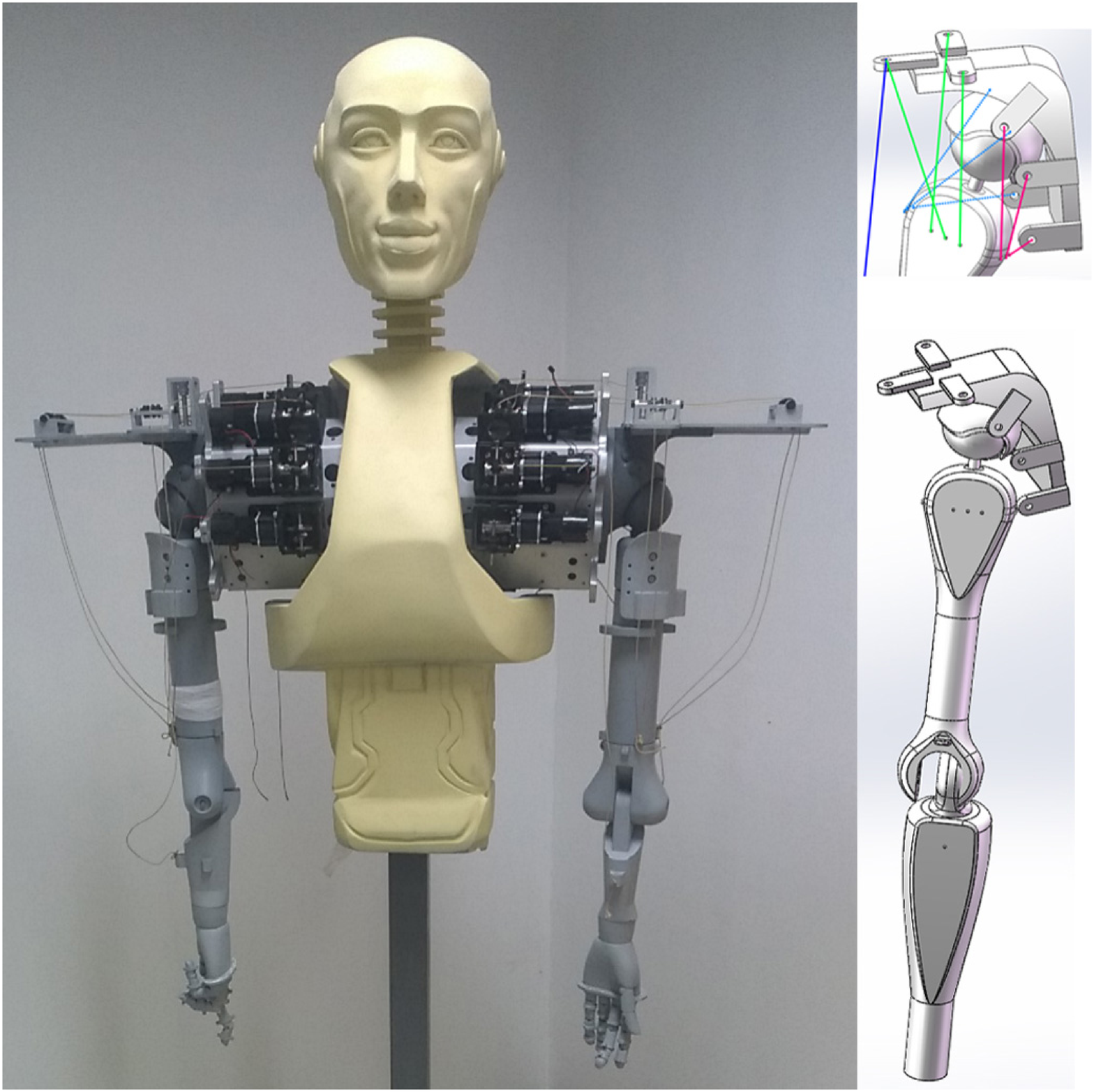}
	\caption{The prototype platform of the neuro-robot, which mimics the muscle-tendon  structure of human movement system.}
	\label{fig:neuro-robot}
\end{figure}

\section*{Acknowledgment}
The authors would like to thank Yongbo Song for his effort on the platform the neuro-robot.

\ifCLASSOPTIONcaptionsoff
  \newpage
\fi


\begin{thebibliography}{10}
	\providecommand{\url}[1]{#1}
	\csname url@samestyle\endcsname
	\providecommand{\newblock}{\relax}
	\providecommand{\bibinfo}[2]{#2}
	\providecommand{\BIBentrySTDinterwordspacing}{\spaceskip=0pt\relax}
	\providecommand{\BIBentryALTinterwordstretchfactor}{4}
	\providecommand{\BIBentryALTinterwordspacing}{\spaceskip=\fontdimen2\font plus
		\BIBentryALTinterwordstretchfactor\fontdimen3\font minus
		\fontdimen4\font\relax}
	\providecommand{\BIBforeignlanguage}[2]{{%
			\expandafter\ifx\csname l@#1\endcsname\relax
			\typeout{** WARNING: IEEEtran.bst: No hyphenation pattern has been}%
			\typeout{** loaded for the language `#1'. Using the pattern for}%
			\typeout{** the default language instead.}%
			\else
			\language=\csname l@#1\endcsname
			\fi
			#2}}
	\providecommand{\BIBdecl}{\relax}
	\BIBdecl
	
	\bibitem{Marques2010}
	H.~G. Marques, M.~J\"{a}ntsch, S.~Wittmeier, O.~Holland, C.~Alessandro,
	A.~Diamond, M.~Lungarella, and R.~Knight, ``Ecce1: the first of a series of
	anthropomimetic musculoskelal upper torsos,'' in \emph{2010 10th IEEE-RAS
		International Conference on Humanoid Robots (Humanoids)}, 2010, pp. 391--396.
	
	\bibitem{Metta2010}
	G.~Metta, L.~Natale, F.~Nori, G.~Sandini, D.~Vernon, L.~Fadiga, C.~von Hofsten,
	K.~Rosander, M.~Lopes, J.~Santos-Victor, A.~Bernardino, and L.~Montesano,
	``The icub humanoid robot: An open-systems platform for research in cognitive
	development,'' \emph{Neural Netw.}, vol.~23, pp. 8--9, 2010.
	
	\bibitem{Fukushima1988}
	K.~Fukushima, ``Neocognitron: a hierarchical neural network capable of visual
	pattern recognition,'' \emph{Neural Netw.}, vol.~1, pp. 119--130, 1988.
	
	\bibitem{Itti1998}
	L.~Itti, C.~Koch, and E.~Niebur, ``A model of saliency-based visual attention
	for rapid scene analysis,'' \emph{IEEE Trans. Pattern Anal. Mach. Intell.},
	vol.~20, no.~11, pp. 1254--1259, 1998.
	
	\bibitem{Itti2006}
	L.~Itti and J.~Bonaiuto, ``The use of attention and spatial information for
	rapid facial recognition in video,'' \emph{Image and Vision Computing},
	vol.~24, no.~6, pp. 557--563, 2006.
	
	\bibitem{Cox2015}
	P.~H. Cox and M.~Riesenhuber, ``There is a "u" in clutter: Evidence for robust
	sparse codes underlying clutter tolerance in human vision.'' \emph{J
		Neurosci.}, vol.~35, no.~42, pp. 14\,148--14\,159, 2015.
	
	\bibitem{Tacchetti2015}
	A.~Tacchetti, L.~Isik, and T.~Poggio, ``Invariant representations for action
	recognition in the visual system,'' \emph{J. Vis.}, vol.~15, no.~12, p. 558,
	2015.
	
	\bibitem{Huang2011}
	Y.~Huang, K.~Huang, D.~Tao, T.~Tan, and X.~Li, ``Enhanced biologically inspired
	model for object recognition,'' \emph{IEEE Trans. Syst., Man, Cybern. B},
	vol.~41, no.~6, pp. 1668--1680, 2011.
	
	\bibitem{Serre2007}
	T.~Serre, L.~Wolf, S.~Bileschi, M.~Riesenhuber, and T.~Poggio, ``Robust object
	recognition with cortex-like mechanisms,'' \emph{IEEE Trans. Pattern Anal.
		Mach. Intell.}, vol.~29, pp. 411--426, 2007.
	
	\bibitem{Theriault2013}
	Th\'{e}riault, N.~C., Thome, and M.~Cord, ``Extended coding and pooling in the
	hmax model,'' \emph{IEEE Trans. Image Process.}, vol.~22, pp. 764--777, 2013.
	
	\bibitem{Qiao2014}
	H.~Qiao, Y.~L. Li, T.~Tang, and P.~Wang, ``Introducing memory and association
	mechanism into a biologically inspired visual model,'' \emph{{IEEE} Trans.
		Syst., Man, Cybern. {B}}, vol.~44, no.~9, pp. 1485--1496, 2014.
	
	\bibitem{Qiao2015}
	H.~Qiao, X.~Y. Xi, Y.~L. Li, W.~Wu, and F.~F. Li, ``Biologically inspired
	visual model with preliminary cognition and active attention adjustment,''
	\emph{{IEEE} Trans. Syst., Man, Cybern. {B}}, vol.~45, no.~11, pp.
	2612--2624, 2015.
	
	\bibitem{Yan2014}
	Z.~Yan, X.~Y. Le, and J.~Wang, ``Model predictive control of linear parameter
	varying systems based on a recurrent neural network.'' in \emph{TPNC}, vol.
	8890.\hskip 1em plus 0.5em minus 0.4em\relax Springer, pp. 255--266.
	
	\bibitem{Lee2009}
	H.~Lee, Grosse, R.~R., R., and A.~Y. Ng, ``Convolutional deep belief networks
	for scalable unsupervised learning of hierarchical representations,'' in
	\emph{ICML}, 2009, pp. 609--616.
	
	\bibitem{Krizhevsky2012}
	A.~Krizhevsky, I.~Sutskever, and G.~E. Hinton, ``Imagenet classification with
	deep convolutional neural networks,'' in \emph{NIPS}, 2012, pp. 1097--1105.
	
	\bibitem{Girshick2014}
	Girshick, D.~R., D.~J., T., and J.~Malik, ``Rich feature hierarchies for
	accurate object detection and semantic segmentation,'' in \emph{CVPR}, 2014,
	pp. 580--587.
	
	\bibitem{Schroff2015}
	F.~Schroff, D.~Kalenichenko, and J.~Philbin, ``Facenet: a unified embedding for
	face recognition and clustering,'' in \emph{CVPR}, 2015, pp. 815--823.
	
	\bibitem{Qiao2016}
	H.~Qiao, Y.~L. Li, F.~F. Li, X.~X. Xi, and W.~Wu, ``Biologically inspired model
	for visual cognition - achieving unsupervised episodic and semantic feature
	learning,'' \emph{{IEEE} Trans. Syst., Man, Cybern. {B}}, vol.
	DOI:10.1109/TCYB.2015.2476706, 2015.
	
	\bibitem{Palmer2015}
	L.~R. Palmer, E.~Diller, and R.~D. Quinn, ``Toward gravity-independent climbing
	using a biologically inspired distributed inward gripping strategy,''
	\emph{{IEEE/ASME} Trans. Mechatronics}, vol.~20, no.~2, pp. 631--640, 2015.
	
	\bibitem{Koo2015}
	I.~M. Koo, T.~D. Trong, Y.~H. Lee, K.~Moon, H., S.~J., Park, and H.~R. Choi,
	``Biologically inspired gait transition control for a quadruped walking
	robot,'' \emph{Auton. Robots}, vol.~39, no.~2, pp. 169--182, 2015.
	
	\bibitem{Srinivasan2006}
	M.~Srinivasan and A.~Ruina, ``Computer optimization of a minimal biped model
	discovers walking and running,'' \emph{Nature}, vol. 439, pp. 72--75, 2006.
	
	\bibitem{Kwon2014}
	J.~Kwon, W.~Yang, H.~Lee, J.-H. Bae, and Y.~Oh, ``Biologically inspired control
	algorithm for an unified motion of whole robotic arm-hand system,'' in
	\emph{RO-MAN}, 2014, pp. 398--404.
	
	\bibitem{Hunt2015}
	A.~Hunt, M.~Schmidt, M.~Fischer, and Q.~R., ``A biologically based neural
	system coordinates the joints and legs of a tetrapod,'' \emph{Bioinspir
		Biomim.}, vol.~10, no.~5, p. 55004, 2015.
	
	\bibitem{Renjewski2015}
	D.~Renjewski, A.~Sprowitz, A.~Peekema, M.~Jones, and J.~Hurst, ``Exciting
	engineered passive dynamics in a bipedal robot,'' \emph{IEEE Trans. Robot.},
	vol.~31, no.~5, pp. 1244--1251, 2015.
	
	\bibitem{Qiao20162}
	H.~Qiao, C.~Li, P.~J. Yin, W.~Wu, and Z.-Y. Liu, ``Human-inspired motion model
	of upper-limb with fast response and learning ability - a promising direction
	for robot system and control,'' \emph{Assembly Automation}, in publication.
	
	\bibitem{Horaud1998}
	R.~Horaud, F.~Dornaika, and B.~Espiau, ``Visually guided object grasping,''
	\emph{IEEE Trans. Robot. Autom.}, vol.~14, no.~4, pp. 525--532, 1998.
	
	\bibitem{Law2014}
	J.~Law, P.~Shaw, M.~Lee, and M.~Sheldon, ``From saccades to grasping: A model
	of coordinated reaching through simulated development on a humanoid robot,''
	\emph{IEEE T. Autonomous Mental Development}, vol.~6, no.~2, pp. 93--109,
	2014.
	
	\bibitem{Lukic2015}
	L.~Lukic, A.~Billard, and J.~Santos-Victor, ``Motor-primed visual attention for
	humanoid robots,'' \emph{IEEE T. Autonomous Mental Development}, vol.~7,
	no.~2, pp. 76--91, 2015.
	
	\bibitem{Lopes2005}
	M.~Lopes and J.~Santos-Victor, ``Visual learning by imitation with motor
	representations,'' \emph{{IEEE} Trans. Syst., Man, Cybern. {B}}, vol.~35,
	no.~3, pp. 438--449, 2005.
	
	\bibitem{Goodale1992}
	M.~A. Goodale and A.~D. Milner, ``Separate visual pathways for perception and
	action,'' \emph{Trends Neurosci.}, vol.~15, no.~1, pp. 20--25, 1992.
	
	\bibitem{Lamme1998}
	V.~Lamme, H.~Supèr, and H.~Spekreijse, ``Feedforward, horizontal, and feedback
	processing in the visual cortex,'' \emph{Curr. Opin. Neurobiol.}, vol.~8,
	no.~4, pp. 529--535, 1998.
	
	\bibitem{Tanaka1993}
	K.~Tanaka, ``Neuronal mechanisms of object recognition,'' \emph{Science}, vol.
	262, pp. 685--688, 1993.
	
	\bibitem{Bear2007}
	M.~Bear, B.~Connors, and M.~Paradiso.\hskip 1em plus 0.5em minus 0.4em\relax MD
	\: Lippincott Williams \& Wilkins., 2007.
	
	\bibitem{deHaan2011}
	E.~H. de~Haan and A.~Cowey, ``On the usefulness of 'what' and 'where' pathways
	in vision,'' \emph{Trends Cogn. Sci.}, vol.~15, no.~10, pp. 460--466, 2011.
	
	\bibitem{Ungerleider1994}
	L.~G. Ungerleider and J.~V. Haxby, ``"what" and "where" in the human brain,''
	\emph{Curr. Opin. Neurobiol.}, vol.~33, no.~4, pp. 157--165, 1994.
	
	\bibitem{Sanes1995}
	J.~N. Sanes, J.~P. Donoghue, V.~Thangaraj, R.~R. Edelman, and S.~Warach,
	``Shared neural substrates controlling hand movements in human motor
	cortex,'' \emph{Science}, vol. 268, no. 5218, pp. 1775--1777, 1995.
	
	\bibitem{Blohm2009}
	G.~Blohm, G.~P. Keith, and J.~D. Crawford, ``Decoding the cortical
	transformations for visually guided reaching in 3d space,'' \emph{Cereb.
		Cortex}, vol.~19, no.~6, pp. 1372--1393, 2009.
	
	\bibitem{Georgopoulos1986}
	A.~P. Georgopoulos, A.~B. Schwartz, and R.~E. Kettner, ``Neuronal population
	coding of movement direction,'' \emph{Science}, vol. 233, no. 4771, pp.
	1416--1419, 1986.
	
	\bibitem{Arce2010}
	F.~Arce, I.~Novick, Y.~Mandelblat-Cerf, Z.~Israel, C.~Ghez, and E.~Vaadia,
	``Combined adaptiveness of specific motor cortical ensembles underlies
	learning,'' \emph{J Neurosci.}, vol.~30, no.~15, pp. 5415--5425, 2010.
	
	\bibitem{Strick2009}
	P.~L. Strick, R.~P. Dum, and J.~A. Fiez, ``Cerebellum and nonmotor function,''
	\emph{Annu. Rev. Neurosci.}, vol.~32, no.~1, pp. 413--434, 2009.
	
	\bibitem{Buckner2013}
	R.~L. Buckner, ``The cerebellum and cognitive function: 25 years of insight
	from anatomy and neuroimaging,'' \emph{Neuron}, vol.~80, no.~3, pp. 807--815,
	2013.
	
	\bibitem{Therrien2015}
	A.~S. Therrien and A.~J. Bastian, ``Cerebellar damage impairs internal
	predictions for sensory and motor function,'' \emph{Curr. Opin. Neurobiol.},
	vol.~33, pp. 127--133, 2015.
	
	\bibitem{Boyden2004}
	E.~S. Boyden, A.~Katoh, and J.~L. Raymond, ``Cerebellum-dependent learning: the
	role of multiple plasticity mechanisms,'' \emph{Annu. Rev. Neurosci.},
	vol.~27, pp. 581--609, 2004.
	
	\bibitem{Brooks2015}
	J.~X. Brooks, J.~Carriot, and K.~E. Cullen, ``Learning to expect the
	unexpected: rapid updating in primate cerebellum during voluntary
	self-motion,'' \emph{Nat. Neurosci.}, vol.~18, no.~9, pp. 1310--1317, 2015.
	
	\bibitem{Bullier2001}
	J.~Bullier, ``Integrated model of visual processing,'' \emph{Brain Res. Brain
		Res. Rev.}, vol.~36, pp. 96--107, 2001.
	
	\bibitem{Klistorner1997}
	A.~Klistorner, C.~D. P., and S.~Crewther, ``Separate magnocellular and
	parvocellular contributions from temporal analysis of the multifocal vep,''
	\emph{Vision Res.}, vol.~37, pp. 2161--2169, 1997.
	
	\bibitem{Treue2003}
	S.~Treue, ``Visual attention: the where, what, how and why of saliency,''
	\emph{Curr Opin Neurobiol.}, vol.~13, no.~4, pp. 428--432, 2003.
	
	\bibitem{Riddoch2010}
	M.~Riddoch, M.~Chechlacz, C.~Mevorach, E.~Mavritsaki, H.~Allen, and G.~W.
	Humphreys, ``The neural mechanisms of visual selection: the view from
	neuropsychology,'' \emph{Ann. N. Y. Acad. Sci.}, vol. 1191, pp. 156--181,
	2010.
	
	\bibitem{Uijlings2013}
	J.~Uijlings, K.~van~de Sande, T.~Gevers, and A.~Smeulders, ``Selective search
	for object recognition,'' \emph{International Journal of Computer Vision},
	vol. 104, no.~2, pp. 154--171, 2013.
	
	\bibitem{Yang1992}
	M.~P. Young, ``Objective analysis of the topological organization of the
	primate cortical visual system,'' \emph{Nature}, vol. 358, no. 6382, pp.
	152--155, 1992.
	
	\bibitem{Li2010}
	N.~Li and J.~J. DiCarlo, ``Unsupervised natural visual experience rapidly
	reshapes size-invariant object representation in inferior temporal cortex,''
	\emph{Neuron}, vol.~67, no.~6, pp. 1062--1075, 2010.
	
	\bibitem{Stryker1991}
	M.~P. Stryker, ``Temporal associations,'' \emph{Nature}, vol. 354, no. 6349,
	pp. 108--109, 1991.
	
	\bibitem{LeCun2006}
	Y.~LeCun, S.~Chopra, R.~Hadsell, M.~Ranzato, and F.~Huang, \emph{A tutorial on
		energy-based learning}.\hskip 1em plus 0.5em minus 0.4em\relax The MIT press,
	2006.
	
	\bibitem{Hinton2002}
	G.~E. Hinton, ``Training products of experts by minimizing contrastive
	divergence,'' \emph{Neural Comput.}, vol.~14, no.~8, pp. 1771--1800, 2002.
	
	\bibitem{Diedrichsen2010}
	J.~Diedrichsen, R.~Shadmehr, and R.~B. Ivry, ``The coordination of movement:
	optimal feedback control and beyond,'' \emph{Trends Cogn. Sci.}, vol.~14,
	no.~1, pp. 31--39, 2010.
	
	\bibitem{DeRugy2012}
	A.~De~Rugy, G.~E. Loeb, and T.~J. Carroll, ``Muscle coordination is habitual
	rather than optimal,'' \emph{J. Neurosci.}, vol.~32, no.~21, pp. 7384--7391,
	2012.
	
	\bibitem{Kandel}
	E.~R. Kandel, J.~H. Schwarz, and T.~M. Jessel.\hskip 1em plus 0.5em minus
	0.4em\relax McGraw-Hill, 2000.
	
	\bibitem{Passot2010}
	P.~Jean-Baptiste, N.~L., and A.~Angelo, ``Internal models in the cerebellum: a
	coupling scheme for online and offline learning in procedural tasks,'' in
	\emph{SAB}, ser. Lecture Notes in Computer Science, vol. 6226, 2010, pp.
	435--446.
	
	\bibitem{Mitsunari2011}
	A.~Mitsunari, S.~Heidi, W.~Eric, N.~S. Dave~L., and C.~Leonardo, ``Reward
	improves long-term retention of a motor memory through induction of offline
	memory gains,'' \emph{Current Biology}, vol.~21, no.~7, pp. 557--562, 2011.
	
	\bibitem{Cantarero2015}
	G.~Cantarero, D.~Spampinato, J.~Reis, L.~Ajagbe, T.~Thompson, K.~Kulkarni, and
	P.~Celnik, ``Cerebellar direct current stimulation enhances on-line motor
	skill acquisition through an effect on accuracy,'' \emph{J. Neurosci.},
	vol.~35, no.~7, pp. 3285--3290, 2015.
	
	\bibitem{Thach1996}
	W.~T. Thach, ``On the specific role of the cerebellum in motor learning and
	cognition: Clues from pet activation and lesion studies in man,''
	\emph{Behavioral and Brain Sciences}, vol.~19, no.~3, pp. 411-- 433, 1996.
	
	\bibitem{Kazuyoushi2004}
	T.~Kazuyoshi and F.~Shintaro, ``Population vector analysis of primate
	prefrontal activity during spatial working memory,'' \emph{Cereb.Cortex},
	vol.~14, no.~12, pp. 1328--1339, 2004.
	
	\bibitem{Jean2007}
	A.~Jean-Marc, H.~Val\'{e}rie, R.~Jean-Pierre, and R.-C. Edith, ``Cutaneous
	afferents provide a neuronal population vector that encodes the orientation
	of human ankle movements,'' \emph{J. Physio.}, vol. 580, no.~2, pp. 649--658,
	2007.
	
	\bibitem{Garner2003}
	B.~A. Garner and M.~G. Pandy, ``Estimation of musculotendon properties in the
	human upper limb,'' \emph{Ann. Biomed. Eng.}, vol.~31, no.~2, pp. 207--220,
	2003.
	
	\bibitem{Davis1999}
	A.~M. Davis, D.~E. Beaton, P.~Hudak, P.~Amadio, C.~Bombardier, D.~Cole,
	G.~Hawker, J.~N. Katz, M.~Makela, R.~G. Marx, L.~Punnett, and J.~G. Wright,
	``Measuring disability of the upper extremity: a rationale supporting the use
	of a regional outcome measure,'' \emph{J. Hand Ther.}, vol.~12, no.~4, pp.
	269--274, 1999.
	
	\bibitem{Thelen2003}
	D.~G. Thelen, F.~C. Anderson, and S.~L. Delp, ``Generating dynamic simulations
	of movement using computed muscle control,'' \emph{J. Biomech.}, vol.~36,
	no.~3, pp. 321--328, 2003.
	
	\bibitem{Thelen2006}
	D.~G. Thelen and F.~C. Anderson, ``Using computed muscle control to generate
	forward dynamic simulations of human walking from experimental data,''
	\emph{J. Biomech.}, vol.~39, no.~6, pp. 1107--1115, 2006.
	
	\bibitem{Pennestri2007}
	E.~Pennestri, R.~Stefanelli, P.~Valentini, and L.~Vita, ``Virtual
	musculo-skeletal model for the biomechanical analysis of the upper limb,''
	\emph{J. Biomech.}, vol.~40, no.~6, pp. 1350--1361, 2007.
	
	\bibitem{Chang2011}
	C.-C. Chang and C.-J. Lin, ``Libsvm: a library for support vector machines,''
	\emph{ACM. T. Intell. Syst. tech.}, vol.~2, no.~3, p.~27, 2011.
	
	\bibitem{Neubeck2006}
	A.~Neubeck and L.~Van~Gool, ``Efficient non-maximum suppression,'' in
	\emph{Proceedings of the 18th International Conference on Pattern Recognition
		- Volume 03}, 2006, pp. 850--855.
	
	\bibitem{Merigan1993}
	W.~H. Merigan and J.~H. Maunsell, ``How parallel are the primate visual
	pathways,'' \emph{Annu. Rev. Neurosci.}, vol.~16, pp. 369--402, 1993.
	
	\bibitem{Lowe1999}
	D.~G. Lowe, ``Object recognition from local scale-invariant features,'' in
	\emph{Proc. of the International Conference on Computer Vision, Corfu}, 1999.
	
	\bibitem{Dalal2005}
	N.~Dalal and B.~Triggs, ``Histograms of oriented gradients for human
	detection.'' in \emph{CVPR}, 2005, pp. 886--893.
	
\end{thebibliography}
\end{document}